\documentclass[runningheads]{llncs}

\usepackage[mobile]{eccv}

\usepackage{eccvabbrv}

\usepackage{graphicx}
\usepackage{booktabs}

\usepackage{url}
\usepackage{graphics} 
\usepackage{multirow} 
\usepackage{mathrsfs} 
\usepackage{amsmath} 

\usepackage[accsupp]{axessibility}  

\usepackage[pagebackref,breaklinks,colorlinks,citecolor=eccvblue]{hyperref}

\usepackage{orcidlink}

\usepackage{amsmath,amsfonts,bm}

\def\1{\bm{1}}

\def\vx{{\bm{x}}}

\DeclareMathAlphabet{\mathsfit}{\encodingdefault}{\sfdefault}{m}{sl}
\SetMathAlphabet{\mathsfit}{bold}{\encodingdefault}{\sfdefault}{bx}{n}

\begin{document}

\title{CRM: Single Image to 3D Textured Mesh with Convolutional Reconstruction Model} 

\titlerunning{Convolutional Reconstruction Model}

\author{Zhengyi Wang\inst{1,3},
Yikai Wang\inst{1},
Yifei Chen\inst{1},
Chendong Xiang\inst{1,3},
Shuo Chen\inst{1},
Dajiang Yu\inst{1},
Chongxuan Li\inst{2},
Hang Su\inst{1} \and
Jun Zhu\thanks{Corresponding author.}\inst{1,3}
}

\authorrunning{Zhengyi Wang et al.}

\institute{
Dept. of Comp. Sci. \& Tech., BNRist Center, Tsinghua-Bosch Joint ML Center, Tsinghua University;\and
Gaoling School of Artificial Intelligence, Renmin University of China, Beijing Key Laboratory of Big Data Management and Analysis Methods, Beijing, China; \and
ShengShu, Beijing, China.
}

\maketitle

\begin{figure}[ht]
	\centering
	\includegraphics[width=0.95\linewidth]{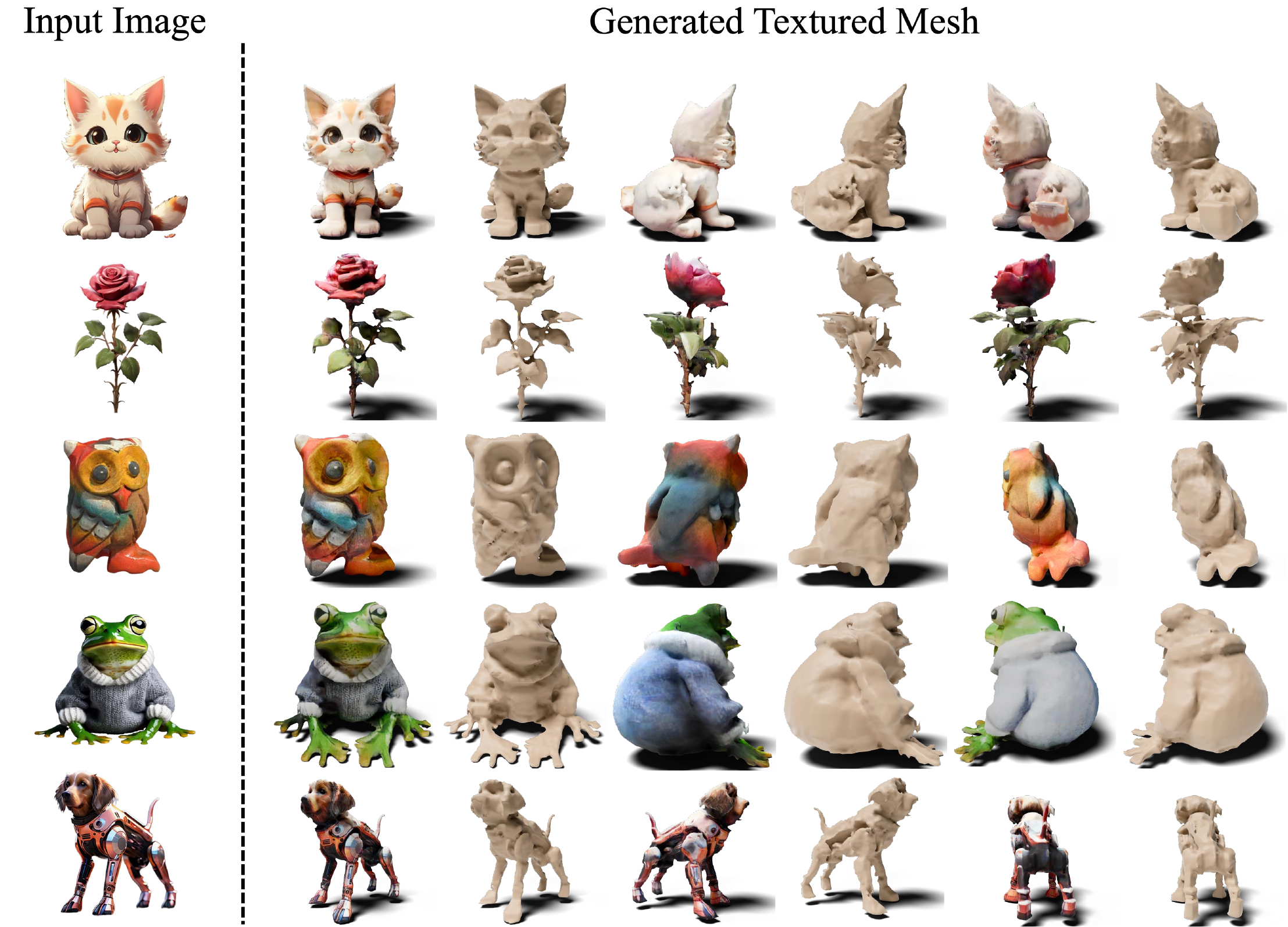}
	\caption{\textbf{CRM} generates high-fidelity textured mesh from single image in \textbf{10 seconds}.}
	\label{fig:head}
\end{figure}
\begin{abstract}

Feed-forward 3D generative models like the Large Reconstruction Model (LRM)~\cite{hong2023lrm} have demonstrated exceptional generation speed. However, the transformer-based methods do not leverage the geometric priors of the triplane component in their architecture, often leading to sub-optimal quality given the limited size of 3D data and slow training. In this work, we present the Convolutional Reconstruction Model (CRM), a high-fidelity feed-forward single image-to-3D generative model. Recognizing the limitations posed by sparse 3D data, we highlight the necessity of integrating geometric priors into network design. CRM builds on the key observation that the visualization of triplane exhibits spatial correspondence of six orthographic images. First, it generates six orthographic view images from a single input image, then feeds these images into a convolutional U-Net, leveraging its strong pixel-level alignment capabilities and significant bandwidth to create a high-resolution triplane. CRM further employs Flexicubes as geometric representation, facilitating direct end-to-end optimization on textured meshes. Overall, our model delivers a high-fidelity textured mesh from an image in just 10 seconds, without any test-time optimization.


\end{abstract}

\section{Introduction}
\label{sec:intro}

In recent years, generative models have witnessed significant advancements, largely attributed to the fast growth in data size. Transformers~\cite{vaswani2023attention}, in particular, have achieved high-performance results across various domains including language~\cite{brown2020language}, image~\cite{bao2022uvit,peebles2023dit} and video generation~\cite{videoworldsimulators2024}. However, the domain of 3D generation presents unique challenges. Unlike the abundance of other modal's data, 3D data is comparatively scarce. The creation of 3D data requires specialized expertise and considerable time, leading to a situation where the largest 3D datasets, namely Objaverse~\cite{deitke2023objaverse,deitke2024objaverse}, only contain millions of 3D content, much smaller than image datasets like Laion~\cite{schuhmann2022laion5b} which contains 5 billion images.

Despite this, recent developments have introduced some transformer-based methods~\cite{hong2023lrm,li2023instant3d,xu2023dmv3d,wang2023pflrm,zou2023triplane} like LRM~\cite{hong2023lrm} for creating 3D content from single or multi-view images in a feed-forward manner. Among these models, the triplane has emerged as a popular component due to its efficiency in generating high-resolution 3D results with minimal memory cost. However, reliance on transformer-based networks for generating triplane patches has not utilized the geometric priors inherent to the triplane concept, leading to sub-optimal results in terms of quality and fidelity, and long training time.

To address the above challenges, this paper presents a new Convolutional Reconstruction Model (CRM) with high generation quality as well as fast training. Given the limited amount of 3D contents, CRM builds on a key hypothesis that it is beneficial to explore geometric priors in architecture design. Namely, we observe from the visualization of triplane~\cite{chen2022tensorf,shi2023zerorf,chen2023single} that triplane exhibits spatial correspondence of input six orthographic images, as shown in Fig.~\ref{fig:triplane-demo}. The silhouette and texture of the input images have a natural alignment with the triplane structure. This motivates us to (1) use six orthographic images as input images to reconstruct 3D contents, which align well as the triplane feature, instead of other arbitrarily chosen poses; (2) use a U-Net convolutional network to map the input images to a rolled-out triplane by exploring the strong pixel-level alignment between the input and output. Furthermore, the significant bandwidth capacity of our U-Net enables a direct transformation of the six orthographic images into the triplane, yielding highly detailed outcomes. Besides, we also add Canonical Coordinate Map (CCM) to the reconstruction network, a novel addition that enriches the model's understanding of spatial relations and geometry.

For the task of 3D generation from a single image, as the six orthographic images and CCMs are not directly available, we train a multi-view diffusion model conditioned on the input image to generate the six orthographic images and another diffusion model to generate the CCMs conditioned on the generated six orthographic images. Both diffusion models are trained on a filtered version of the Objaverse dataset~\cite{deitke2023objaverse}. To further enhance quality and robustness, we implement training improvements for the multi-view diffusion models, including Zero-SNR~\cite{lin2024common}, random resizing, and contour augmentation.

Finally, as directly optimizing high-quality textured meshes is challenging, we adopt Flexicubes~\cite{shen2023flexible} as the geometry representation to falicitate gradient-based mesh optimization. This is unlike previous works~\cite{hong2023lrm,zou2023triplane} that use alternative 3D representation like NeRF~\cite{mildenhall2021nerf} or Gaussian Splatting~\cite{kerbl20233d}. Such methods often involve extra procedure steps to obtain textured meshes~\cite{tang2024lgm}, although they can produce detailed visualizations. With our designs, we are able to train CRM with textured mesh as the final output in an end-to-end manner, and our approach has a more straightforward inference pipeline and better mesh quality.
Overall, our method can generate high-fidelity textured mesh within 10 seconds, as shown in Fig.~\ref{fig:head}.

\section{Related Works}
\label{sec:related}
\subsection{Score Distillation for 3D Generation}
DreamFusion~\cite{poole2022dreamfusion} proposes a technique called Score Distillation Sampling (SDS) (also known as Score Jacobian Chaining~\cite{wang2022sjc}). It utilizes large scale image diffusion models~\cite{saharia2022photorealistic,rombach2022high} to iteratively refine 3D models to align with specific prompts or images. Thus it can generate 3D content without training on 3D dataset. Along this line, ProlificDreamer~\cite{wang2023prolificdreamer} proposes Variational Score Distillation (VSD), a principled variational framework which greatly mitigates the over-saturation problems in SDS and improves the diversity. Zero123~\cite{liu2023zero123}, MVDream~\cite{shi2023mvdream}, ImageDream~\cite{wang2023imagedream} and many others~\cite{qian2023magic123,li2023sweetdreamer,qiu2023richdreamer} further improve the results and mitigate the multi-face problems using diffusion models fine-tuned on 3D data. \cite{lorraine2023att3d,qian2024atom} explore amortized score distillation. Many other works~\cite{lin2023magic3d,chen2023fantasia3d,zhu2023hifa,tsalicoglou2023textmesh,chen2023it3d,tang2023dreamgaussian,chen2023gsgen,liang2023luciddreamer,sun2023dreamcraft3d,yu2023csd,liu2023sherpa3d,kim2023collaborative,decatur2023paintbrush,wang2023animatabledreamer,li2023focaldreamer} improve the results a lot, in either speed or quality. However, methods based on score distillation usually take from minutes to hours to generate single object, which is computationally expensive.

\subsection{3D Generation with Sparse View Reconstruction}
Several approaches aim to generate multi-view consistent images and then create 3D contents using sparse views reconstruction. For example, SyncDreamer~\cite{liu2023syncdreamer} generates multi-view consistent images and then uses NeuS~\cite{wang2023neus} for reconstruction. Wonder3D~\cite{long2023wonder3d} improves the results with cross-domain diffusion. Direct2.5~\cite{lu2023direct25} improves the results with 2.5D diffusion. However, one common issue of these methods is that they need test-time optimization for reconstruction with sparse views, which may lead to extra computing and compromise the final quality

\subsection{Feed-forward 3D Generative Models}
Some works try to generate 3D objects using a feed forward model~\cite{chan2022efficient,gao2022get3d,gupta20233dgen,zheng2023las,cheng2023sdfusion,zeng2022lion}. Feed-forward methods demonstrate significantly faster generation speeds compared to the two types of methods mentioned above. Recently there are some works trained on larger 3D dataset Objaverse~\cite{deitke2023objaverse}. One-2-3-45~\cite{liu2023one} generates multi-view images and then feed the images into a network to get the 3D object. LRM series~\cite{hong2023lrm,li2023instant3d,xu2023dmv3d,wang2023pflrm} improve the quality of generated results with a transformer-based architecture. TGS~\cite{zou2023triplane} and LGM~\cite{tang2024lgm} use Gaussian Splatting~\cite{kerbl20233d} as the geometry representation. There are also many other works~\cite{liu2023onepp,zheng2024mvd2,TripoSR2024} that improve the results with different techniques. Despite these advancements, there remains room for improvement in the network architecture or geometry representation. Our approach utilizes a network with a strategically designed architecture and an end-to-end training approach producing meshes directly as the final output.

\begin{figure}[tb]
  \centering
  \includegraphics[width=0.99\linewidth]{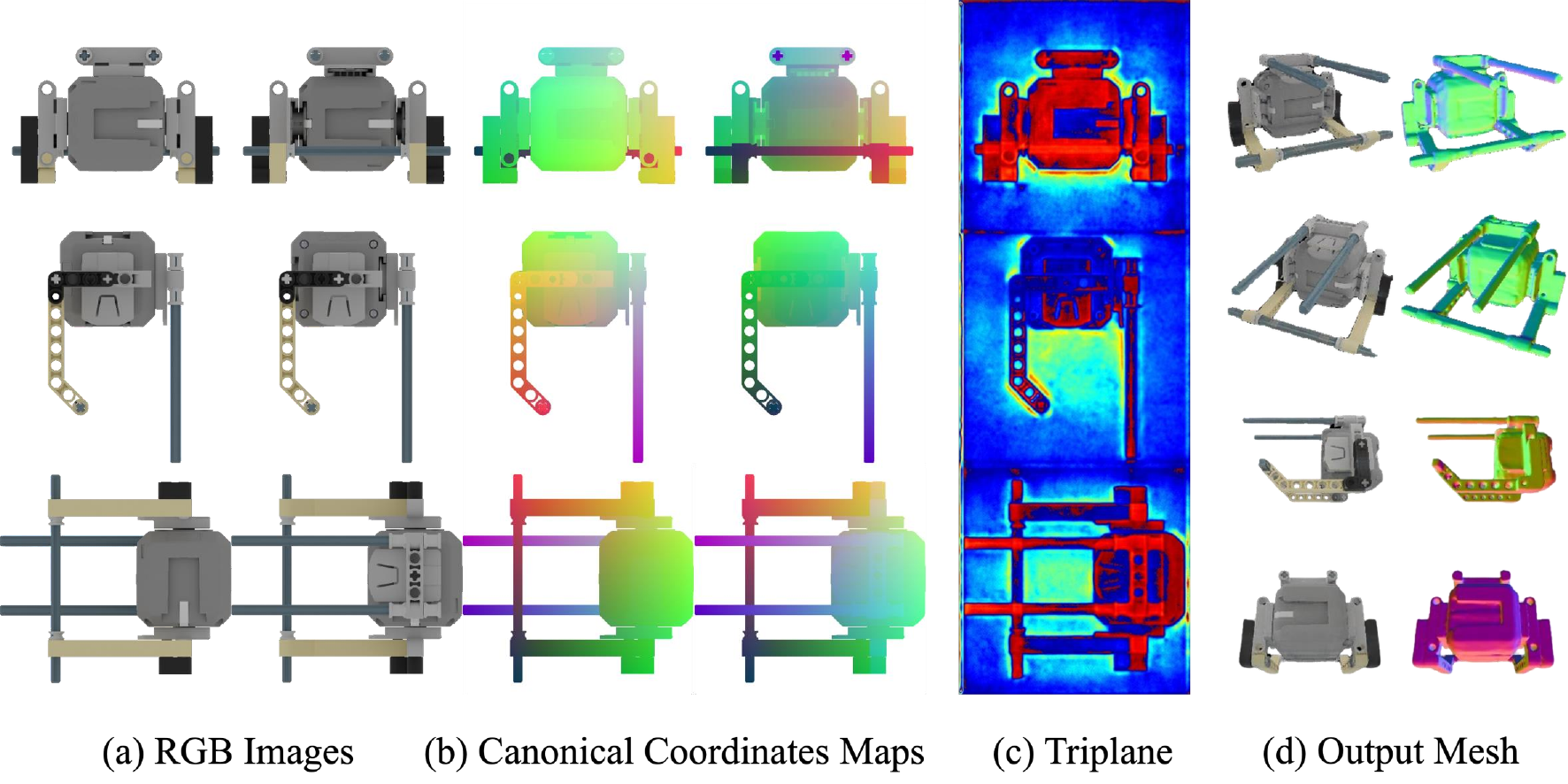}
  \caption{One of our key motivation is that the triplane shares a strong spatial alignment with the input six orthographic images. (a) The six orthographic images of a input shape. (b) The six orthographic CCMs. (c) The triplane (mean value of all channels) output by our U-Net, which spatially aligns with the input images. (d) The textured mesh output by our convolutional reconstruction model.
  }
  \label{fig:triplane-demo}
\end{figure}

\section{Method}

\begin{figure}[tb]
  \centering
  \includegraphics[width=0.99\linewidth]{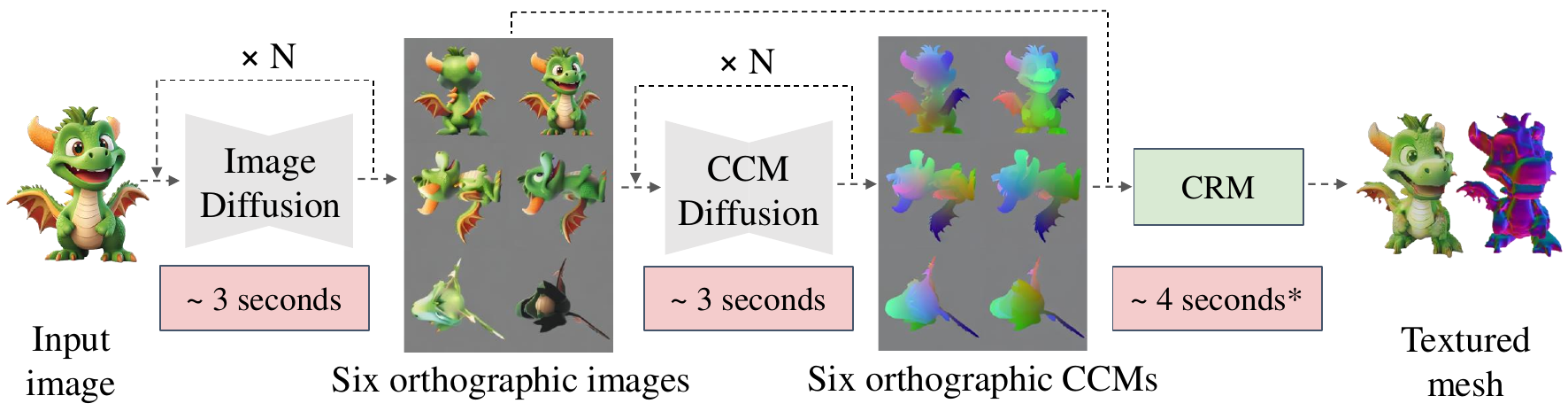}
  \caption{Overall pipeline of our method. The input image is fed into a multi-view image diffusion model to generate six orthographic images. Then another diffusion model is used to generate the CCMs conditioned on the six images. The six images along with the CCMs are send into CRM to reconstruct the final textured mesh. The whole inference process takes around 10 seconds on an A800 GPU. *The 4 seconds includes the U-Net forward (less than $0.1$s), querying surface points for UV texture and file I/O.
  }
  \label{fig:inference}
\end{figure}

In this section, we illustrate the detailed design of our method (shown in Fig.~\ref{fig:inference}). Given a single input image, our model first utilizes multi-view diffusion models (Sec.~\ref{sec:mvdiffusion}) to generate six orthographic images and the canonical coordinates maps (CCMs).
Then we develop our \textbf{convolutional reconstruction model} (CRM, Sec.~\ref{sec:crm}) to reconstruct 3D textured mesh from the images and CCMs.

\label{sec:method}

\subsection{Multi-view Diffusion Model}
\label{sec:mvdiffusion}
We first explain the design of the multi-view diffusion model to generate six orthographic view images from a single input image. Instead of training from scratch, which is typically extremely expensive, we initialize the diffusion models using the checkpoint of ImageDream~\cite{wang2023imagedream}, a high-performance diffusion model for the task of multi-view images generation from a single image. The original ImageDream supports $4$ views generation. We expand it to include 6 views by adding two more perspectives (up and down).
We use another diffusion model which is conditioned on the generated six views to generate the canonical coordinate map. The conditional RGB image is concatenated with the noisy canonical coordinate map. It is also initialized from ImageDream checkpoint. Both the diffusion models are fine-tuned on the Objaverse~\cite{deitke2023objaverse} dataset.

To further improve the quality and robustness of our results, we introduce several enhancements: \textbf{(1) Zero-SNR Training.} We use the zero-SNR trick as mentioned in~\cite{lin2024common}. This can alleviate the problem resulting from the discrepancy between the initial Gaussian noise during sampling and the noisiest training sample. \textbf{(2) Random Resizing.} A naive implementation would make the model tends to generate objects that occupy the entire image. To mitigate this, we randomly resize the objects when training. \textbf{(3) Contour Augmentation.} We find that the model tends to predict the backview color largely relying on the contour of the input view. To make the model insensitive to the contour, we randomly change the contour color during training. 

\subsection{Convolutional Reconstruction Model}
\label{sec:crm}
\begin{figure}[tb]
  \centering
  \includegraphics[width=0.99\linewidth]{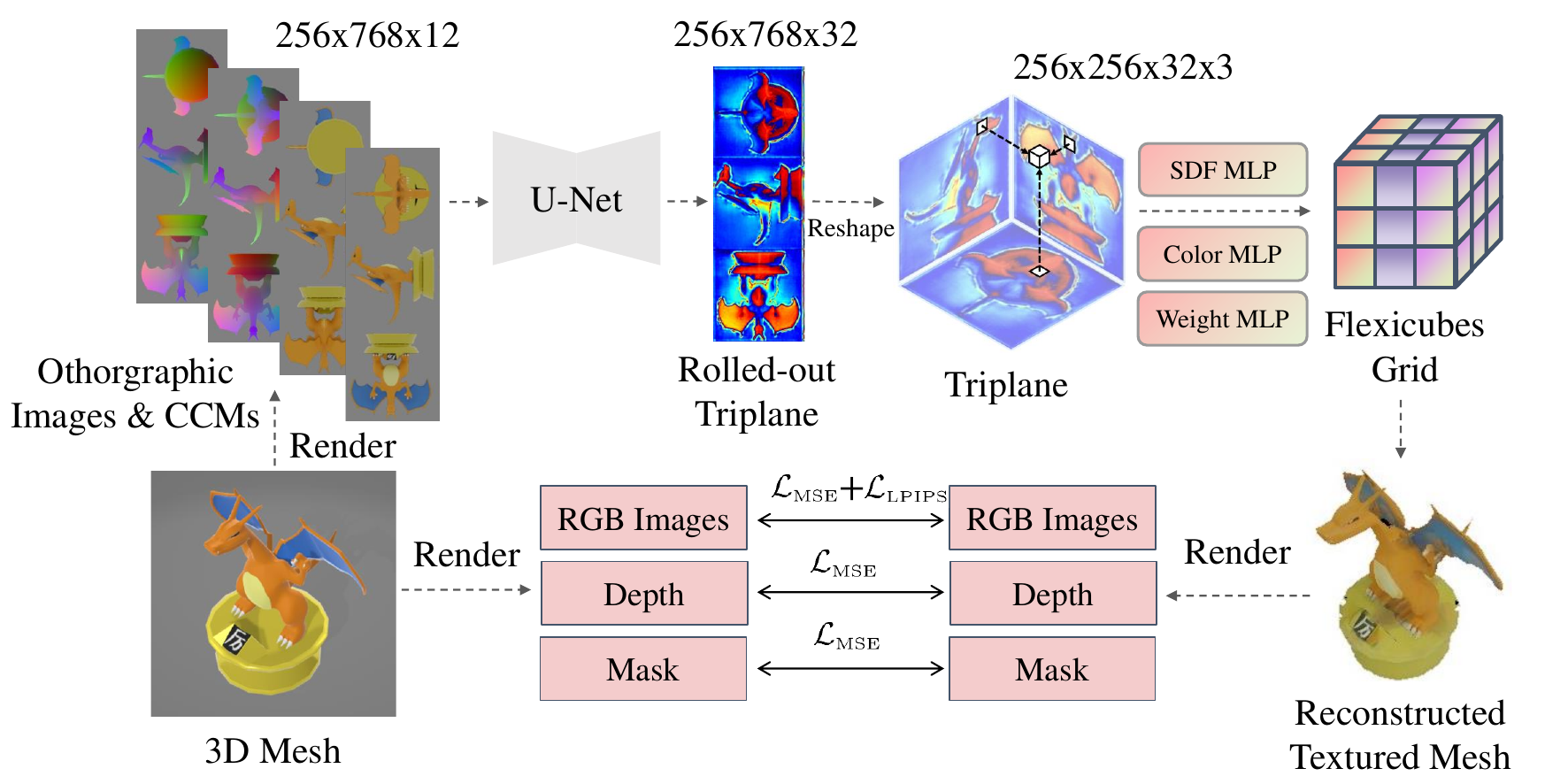}
  \caption{Architecture along with training pipeline of CRM. We render the 3D mesh into six orthographic images and CCMs. Then the images and CCMs are concatenated and fed into the U-Net. The output triplane is decoded by small MLP networks to form the feature grid of Flexicubes, then textured mesh is get by dual marching cubes. During training, we render the color images, depth maps and masks from GT mesh and reconstructed mesh for supervision.
  }
  \label{fig:diagram-all}
\end{figure}

We now move to introduce the detailed architecture of the convolutional reconstruction model (CRM). As outlined in Fig.~\ref{fig:diagram-all}, given the input six images and CCMs, a convolutional U-Net is used to map the input images along with the CCMs to a rolled-out triplane. Then the rolled-out triplane is reshaped into the triplane. Small multi-layer perceptions (MLPs) are used to decode the triplane features into SDF values, texture color and Flexicubes parameters. Lastly, these values are used to get texture mesh by dual marching cubes. Below, we explain the key components of CRM in detail.

\subsubsection{Triplane Representation}

We choose triplane as the 3D representation, because it can achieve high resolution 3D results with 2D computation consumption.  It projects each query grid cell to axis-aligned orthogonal planes ($xy$, $xz$, and $yz$ planes) and then aggregates the feature from each planes. Then the feature is decoded by 3 tiny MLPs with 2 hidden layers to get the SDF values along with deformation, color and Flexicubes weights, respectively. Further, to avoid the unnecessary entanglements of different planes, we use rolled-out triplane~\cite{wang2023rodin}.

\subsubsection{Canonical Coordinates Map (CCM)}

We also add CCM as input~\cite{li2023sweetdreamer}, which contains extra geometry information. This is different from previous works, which typically use pure RGB images as the input to predict the 3D object~\cite{hong2023lrm}. Using pure RGB images make it extremely hard to predict the correct geometry, and sometime the geometry degrades (details in Sec.~\ref{sec:ablation}). Formally, CCM is the coordinates of each point in canonical space. It contains 3 channels whose values are within $[0, 1]$, representing the coordinates in the canonical space.

\subsubsection{UNet-based Convolutional Network}
Our key insight is that the triplane is spatially aligned with the input six orthographic images and CCMs, as shown in Fig.~\ref{fig:triplane-demo}. To match the rolled-out triplane, the six images and CCMs are arranged in a similar way. We render the six images and CCMs at a resolution of $256\times 256$. They are split into two groups, with each group holding three images. These images in the four groups are then combined to create four larger images, each with a resolution of $256\times 768$, allowing for spatial alignment. By concatenating these four groups, we form a 12-channel input. Next, a convolutional U-Net processes this input to produce the output triplane.

Compared to transformer-based methods~\cite{hong2023lrm,li2023instant3d,xu2023dmv3d,zou2023triplane}, our U-shape design has a larger bandwidth in preserving the input information, leading to highly detailed triplane features and finally elaborate textured meshes. Moreover, the convolutional network fully utilizes the geometry prior of the spatial correspondance of triplanes and input six orthograhic images, which greatly fasten the convergence and stabilize the training. Our model can get reasonable reconstruction results at very early stage of training (around 20 minutes of training from scratch). Also, our model can be trained with a much smaller batch size $32$ (compared to transformer-based LRM that uses a batch size of $1024$), which makes that all of our experiments can be conducted on an 8-GPU-cards machine. The overall training cost of our reconstruction model is only $1/8$ than LRM. More details are shown in the experiments (see Sec.~\ref{sec:exp-setting}).

\subsubsection{Flexicubes Geometry} Previous generic 3D generation methods mostly adopt NeRF~\cite{mildenhall2021nerf} or Gaussian splatting~\cite{kerbl20233d} as the geometry representation, which relies on extra procedures like Marching Cubes (MC) 
to extract the iso-surface, suffering from topological ambiguities and struggling to represent high-fidelity geometric details. In this work, we use Flexicubes~\cite{shen2023flexible} as our geometry representation. It can get meshes from the features on the grid by dual marching cubes~\cite{nielson2004dual} during training. The features include SDF values, deformation and weights. The texture is obtained by querying the color at the surface. Flexicubes enables us to train our reconstruction model with textured mesh as the final output in an end-to-end manner.

\subsubsection{Loss Function} Finally, to train our CRM model, we use a combination of MSE loss $\mathcal{L}_{\mbox{\tiny MSE}}$ and LPIPS loss~\cite{zhang2018unreasonable} $\mathcal{L}_{\mbox{\tiny LPIPS}}$ for the rendered images for texture, similar as LRM~\cite{hong2023lrm}. To further enhance the geometry, we also include the depth map and mask for supervision~\cite{gupta20233dgen}. The overall loss function is
\begin{equation}
\begin{aligned}
    \mathcal{L} &= \mathcal{L}_{\mbox{\tiny MSE}}(\vx, \vx^{\mbox{\tiny GT}})+\lambda_{\mbox{\tiny LPIPS}} \mathcal{L}_{\mbox{\tiny LPIPS}}(\vx, \vx^{\mbox{\tiny GT}})\\
    &+\lambda_{\mbox{\tiny depth}}\mathcal{L}_{\mbox{\tiny MSE}}(\vx_{\mbox{\tiny depth}},\vx_{\mbox{\tiny depth}}^{\mbox{\tiny GT}})+\lambda_{\mbox{\tiny mask}}\mathcal{L}_{\mbox{\tiny MSE}}(\vx_{\mbox{\tiny mask}},\vx_{\mbox{\tiny mask}}^{\mbox{\tiny GT}})+\lambda_{\mbox{\tiny reg}}\mathcal{L}_{\mbox{\tiny reg}}\mbox{,}
\end{aligned}
\end{equation}
where $\vx$, $\vx_{\mbox{\tiny depth}}$ and $\vx_{\mbox{\tiny sil}}$ represent the RGB image, depth map and mask rendered from the reconstruction textured mesh, respectively. And   $\vx^{\mbox{\tiny GT}}$, $\vx_{\mbox{\tiny depth}}^{\mbox{\tiny GT}}$ and $\vx_{\mbox{\tiny mask}}^{\mbox{\tiny GT}}$ are rendered from the ground truth textured mesh. $\mathcal{L}_{\mbox{\tiny reg}}$ is the mesh-quality regularizers introduced in Flexicubes~\cite{shen2023flexible}. $\lambda_{\mbox{\tiny LPIPS}}$, $\lambda_{\mbox{\tiny depth}}$, $\lambda_{\mbox{\tiny mask}}$ and $\lambda_{\mbox{\tiny reg}}$ are the coefficients that balance each loss.

\begin{figure}[!t]
	\centering
	\includegraphics[width=0.99\linewidth]{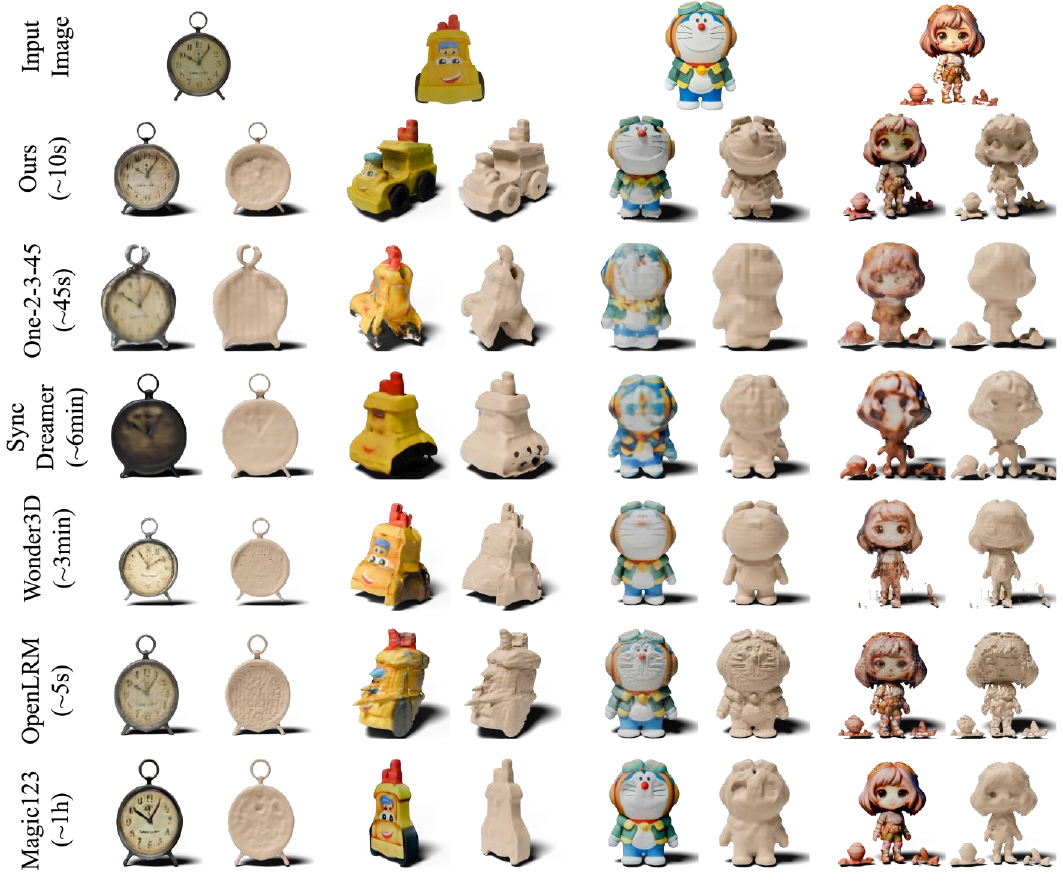}
	\caption{Qualitative comparison with baselines. Our models generates high-fidelity results with better geometry and texture.}
	\label{fig:compare1}
\end{figure}

\section{Experiments}
\label{sec:exp}
\subsection{Experimental Setting}
\label{sec:exp-setting}
\subsubsection{Dataset} We filter the Objaverse~\cite{deitke2023objaverse} dataset, removing scene-level objects and low quality meshes, and get around 376k valid high quality objects as the training set. We reuse the rendered images from SyncDreamer~\cite{liu2023syncdreamer} which contain $16$ images per shape at the resolution of $256\times 256$, and additionally render 6 orthographic images and CCM with the same lighting and resolution.

\subsubsection{Network Architecture} The reconstruction model contains around 300M parameters. The U-Net contains $[64,128,128,256,256,512,512]$ channels, with attention blocks at resolution $[32,16,8]$. We set the Flexicubes grid size as $80$.

\subsubsection{Implementation Details} The reconstruction model was trained on 8 NVIDIA A800 80GB GPU cards for 6 days with $110k$ iterations. The model was trained with batch size $32$ ($32$ shapes per iteration). At each iteration, we randomly sampled $8$ views among the total $16$ images for each shape for supervision. We used the Adam optimizer with learning rate $1e-4$. The coefficients that balancing each loss were set as $\lambda_{\mbox{\tiny LPIPS}}=0.1$, $\lambda_{\mbox{\tiny depth}}=0.5$, $\lambda_{\mbox{\tiny mask}}=0.5$ and $\lambda_{\mbox{\tiny reg}}=0.005$. To enhance robustness against minor inconsistency in the generated multi-view images, we introduced small Gaussian noise to the inputs in both training and inference.

The diffusion models for both six orthographic images and CCMs were trained on 8 NVIDIA A800 80GB GPU cards for 2 days with $10k$ iterations. The gradient accumulation is set as $12$ steps, yielding a total batch size of $1536$. We used the Adam optimizer with learning rate $5e-5$. During sampling both diffusions were sampled with 50 steps using DDIM~\cite{song2020denoising}.

\subsection{Comparison with baselines}
\subsubsection{Qualitative Results}
To validate the effectiveness of our method, we qualitatively compare our results with previous works including Wonder3d~\cite{long2023wonder3d}, SyncDreamer~\cite{liu2023syncdreamer}, Magic123~\cite{qian2023magic123}, One-2-3-45~\cite{liu2023one} and OpenLRM~\cite{openlrm}. Since LRM~\cite{hong2023lrm} is not open-sourced, we use OpenLRM~\cite{openlrm}, an open-sourced implementation of LRM for comparisons. For the other baselines, we use their official codes and checkpoints. As for the input images for testing, we choose two from GSO~\cite{downs2022google} dataset, one downloaded from web and one generated by text-to-image diffusion model. The results are shown in Fig.~\ref{fig:compare1}. It can be seen from the figure that our method generates 3D textured meshes with better texture and geometry than all other baselines. This is because our reconstruction model fully utilizes the spatial alignment of input six orthographic images and output triplane. Also, our model can generate with only 10 seconds, much faster than most of the baselines. Our method is trained in an end-to-end manner with textured mesh as final output, thus avoiding a time-consuming post-processing for converting to mesh as in~\cite{tang2024lgm}.

\begin{figure}[th]
	\centering
	\includegraphics[width=0.99\linewidth]{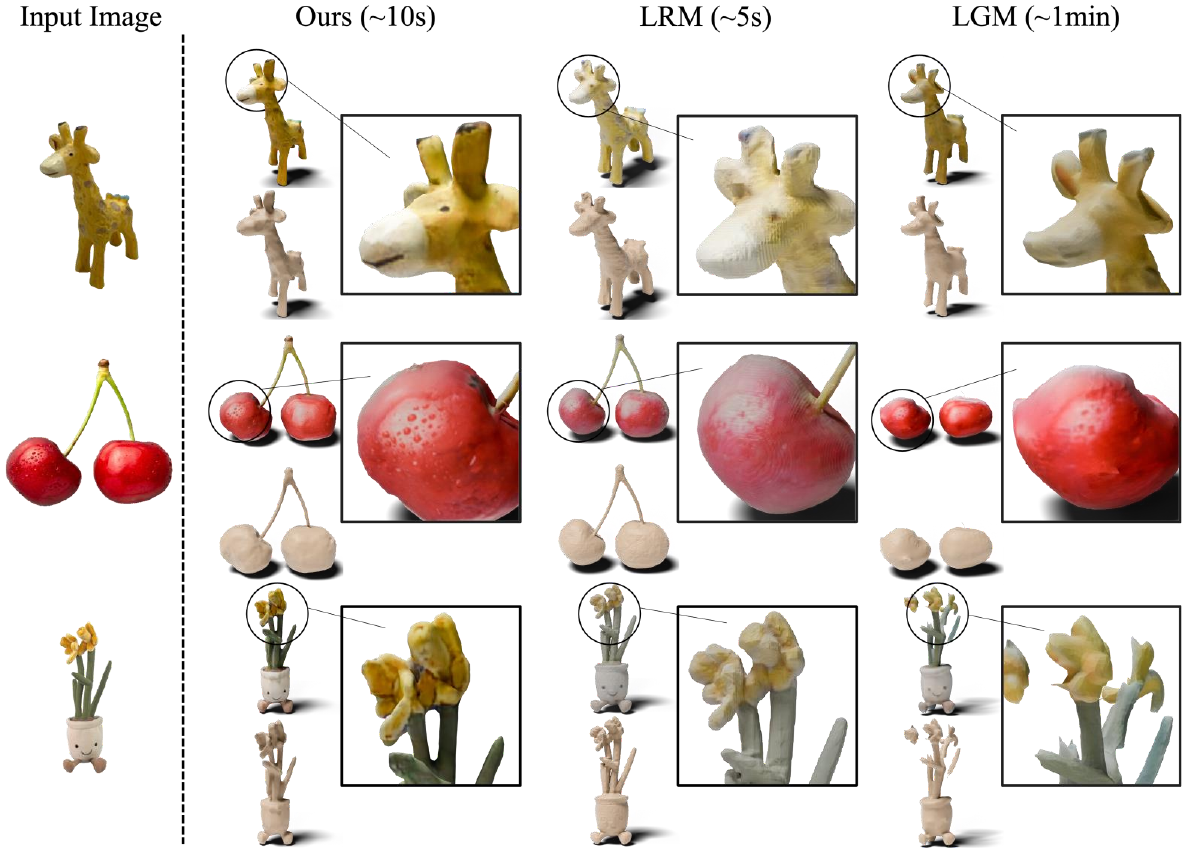}
	\caption{Qualitative comparison with LRM~\cite{hong2023lrm} and LGM~\cite{tang2024lgm}. Our models generates high-fidelity results with detailed texture and smooth geometry.}
	\label{fig:compare2}
\end{figure}

Additionally, we visualize the generated meshes comparing to the previous work LRM~\cite{hong2023lrm} and a concurrent work LGM~\cite{tang2024lgm}. Since LRM is not open-sourced, we use the meshes from their project page. The results are shown in Fig.~\ref{fig:compare2}. It can be seen from the figure that our results have better texture. Our method also has smoother geometry than LRM and better geometry details than LGM.

\begin{figure}[!ht]
	\centering
	\includegraphics[width=0.99\linewidth]{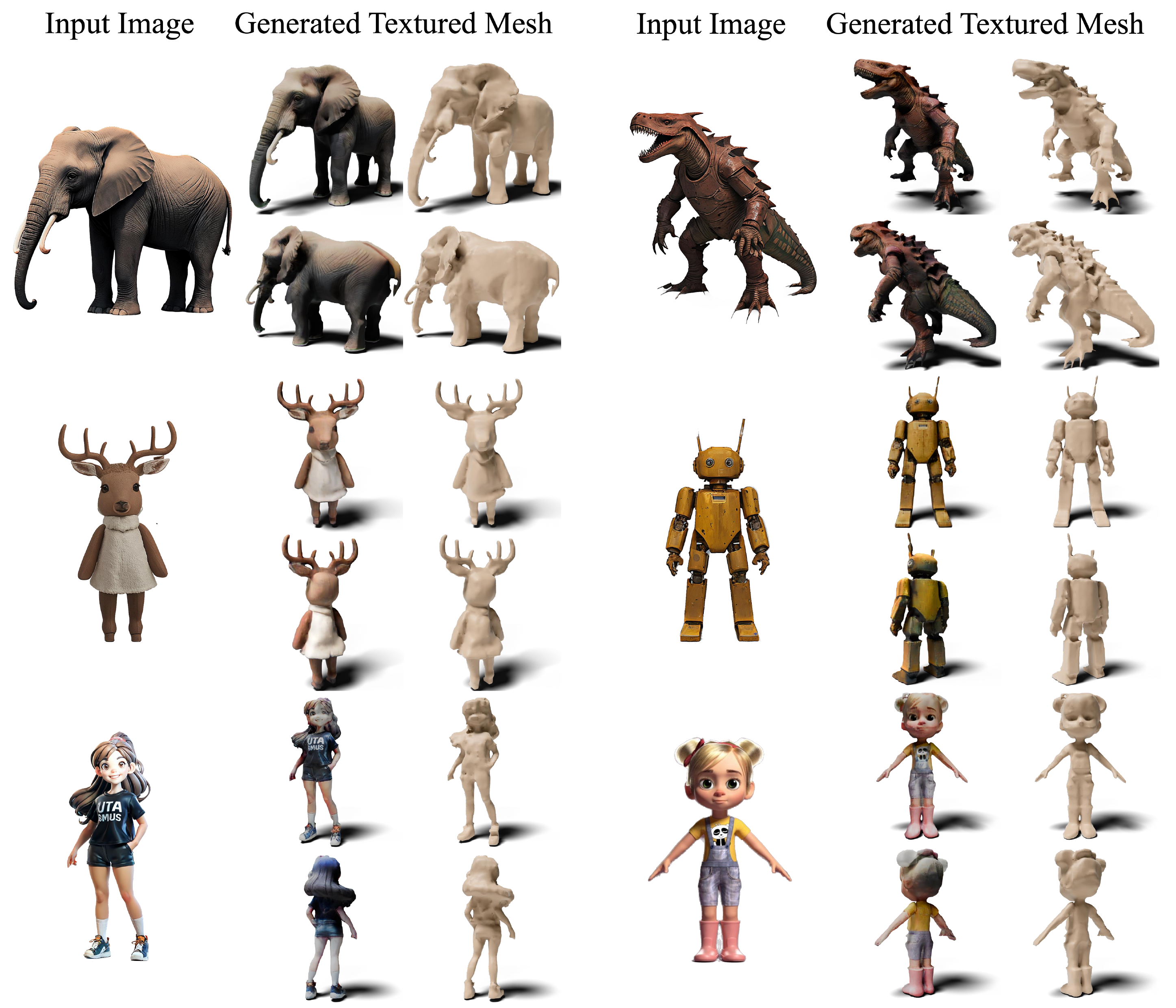}
	\caption{More Results of the generated results of our method from a single image.}
	\label{fig:compare3}
\end{figure}

In Fig.~\ref{fig:compare3}, we show more results of the high-fidelity textured meshes generated from single image by our method.

\subsubsection{Quantitative Results}

\begin{table}[tb]
  \caption{Quantitative comparison for the geometry quality between our method and baselines for single image to 3D textured mesh generation. We report the metrics of Chamfer Distance, Volumn IoU and F-score on GSO dataset.
  }
  \label{tab:quant1}
  \centering
  \begin{tabular}{@{}lccc@{}}
    \toprule
    Method & Chamfer Dist.$\downarrow$& Vol. IoU$\uparrow$ & F-Sco. (\%)$\uparrow$\\
    \midrule
    One-2-3-45~\cite{liu2023one}    & 0.0172 & 0.4463& 72.19\\
    SyncDreamer~\cite{liu2023syncdreamer}    & 0.0140 & 0.3900& 75.74\\
    Wonder3D~\cite{long2023wonder3d}    & 0.0186 & 0.4398&76.75\\
    Magic123~\cite{qian2023magic123} & 0.0188 & 0.3714& 60.66\\
    TGS~\cite{zou2023triplane}    & 0.0172  &0.2982&65.17\\
    OpenLRM~\cite{hong2023lrm, openlrm}    &0.0168   & 0.3774&63.22\\
    LGM~\cite{tang2024lgm} & 0.0117& 0.4685&68.69\\
    Ours    & \textbf{0.0094}  &\textbf{0.6131}& \textbf{79.38}\\
  \bottomrule
  \end{tabular}
\end{table}

\begin{table}[tb]
  \caption{Quantitative comparison for the texture quality between our method and baselines for single image to 3D textured mesh generation. We report the metric of PSNR, SSIM, LPIPS and Clip~\cite{radford2021learning}-Similarity on GSO dataset.
  }
  \label{tab:quant3}
  \centering
  \begin{tabular}{@{}lccccc@{}}
    \toprule
    Method & PSNR$\uparrow$$\mbox{  }$& SSIM$\uparrow$$\mbox{  }$& LPIPS$\downarrow$$\mbox{  }$ & Clip-Sim$\uparrow$ \\
    \midrule
    One-2-3-45~\cite{liu2023one} & 13.93  & 0.8084 & 0.2625 & 79.83\\
    SyncDreamer~\cite{liu2023syncdreamer}  & 14.00  & 0.8165 & 0.2591 & 82.76 \\
    Wonder3D~\cite{long2023wonder3d} & 13.31 & 0.8121 &0.2554 & 83.70 \\
    Magic123~\cite{qian2023magic123} & 12.69 & 0.7984 & 0.2442 & 85.16 \\ 
    OpenLRM~\cite{hong2023lrm,openlrm}    & 14.30  & 0.8294 & 0.2276 & 84.20  \\
    LGM~\cite{zou2023triplane}     & 13.28  &0.7946 & 0.2560 & 85.20 \\
    Ours    & \textbf{16.22}  & \textbf{0.8381} & \textbf{0.2143} & \textbf{87.55}  \\
  \bottomrule
  \end{tabular}
\end{table}

\begin{table}[tb]
  \caption{Quantitative comparison between our method and baselines for novel view synthesis of the multi-view diffusion model. We report the metric of PSNR, SSIM and LPIPS on GSO dataset.
  }
  \label{tab:quant2}
  \centering
  \begin{tabular}{@{}lccc@{}}
    \toprule
    Method & PSNR$\uparrow$$\mbox{ }$& SSIM$\uparrow$$\mbox{ }$& LPIPS$\downarrow$ \\
    \midrule
    SyncDreamer~\cite{liu2023syncdreamer}  &  20.30 & 0.7804 & 0.2932 \\
    Wonder3D~\cite{long2023wonder3d} & 23.76 & 0.8127 & 0.2210\\
    Ours & \textbf{29.36}&\textbf{0.8721}& \textbf{0.1354}\\
  \bottomrule
  \end{tabular}
\end{table}

In line with previous studies~\cite{long2023wonder3d}, we evaluate our method using the Google Scanned Objects (GSO) dataset~\cite{downs2022google} which is not included in our training dataset. We randomly choose 30 shapes and render  a single image with size of $256\times 256$ as input for evaluation. To ensure the generated mesh accurately aligned with the ground truth mesh, we carefully adjust their pose and scale them to fit within the $[-0.5, 0.5]$ box. For mesh geometry evaluation, we report Chamfer Distance (CD), Volumn IoU and F-Score (with a threshold of 0.05, following One-2-3-45~\cite{liu2023one}), which measure the geometry similarity between the reconstructed mesh and ground truth mesh. The results are shown in Table~\ref{tab:quant1}. It can be seen from the table that our method outperforms all of the baselines, which demonstrates the effectiveness of our method for geometry quality.

Furthermore, for evaluating mesh texture, we render 24 images at $512\times 512$ resolution at elevation angles of $0$, $15$, and $30$ degrees, for the generated meshes and ground-truth meshes respectively. For each elevation, the 8 images are evenly distributed around a full 360-degree rotation. Then we assessed them using several metrics: PSNR, SSIM, LPIPS and Clip-Similarity, which measure the resemblance in appearance between the reconstructed mesh and the original ground truth mesh. The results are shown in Table~\ref{tab:quant3}. It shows that the generated textured meshes by our model surpass those of all baselines in appearance, which demonstrates the effectiveness of our method for texture quality. 

Additionally, we carry out experiments to evaluate the effectiveness of our single image to multi-view diffusion models. We use PSNR, SSIM, and LPIPS to measure the similarity between the generated multi-view images and ground truth multi-view images. For this analysis, we use four views of the generated images (left, right, front, back) from each model under comparison, setting the background color to grey (value 128). We compare with SyncDreamer~\cite{liu2023syncdreamer} and Wonder3D~\cite{long2023wonder3d}. The outcomes of this evaluation are documented in Table~\ref{tab:quant2}. It can be seen from the table that our method outperforms all the baselines.

\subsection{Ablation Study and Analysis}
\label{sec:ablation}

\begin{figure}
    \centering
    \begin{minipage}{.45\textwidth}
	\centering
	\includegraphics[height=0.9\linewidth]{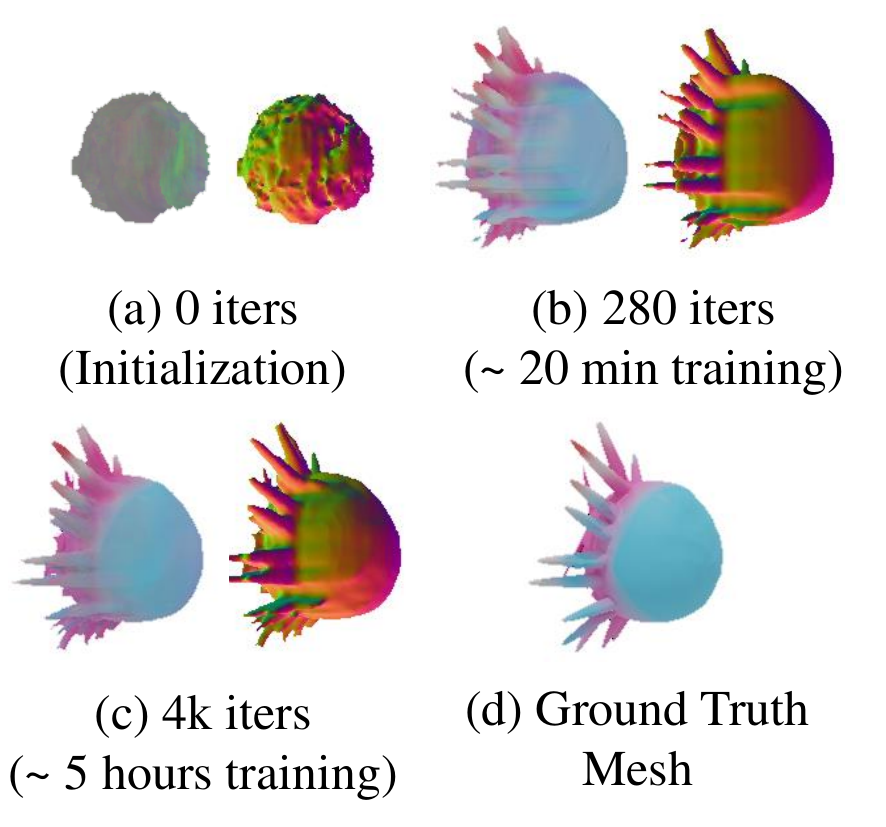}
	\caption{Reconstruction results on unseen samples during early stage of training.}
	\label{fig:ablation-training}
    \end{minipage}
    \hspace{1.5em}
    \begin{minipage}{.45\textwidth}
	\centering
	\includegraphics[height=0.9\linewidth]{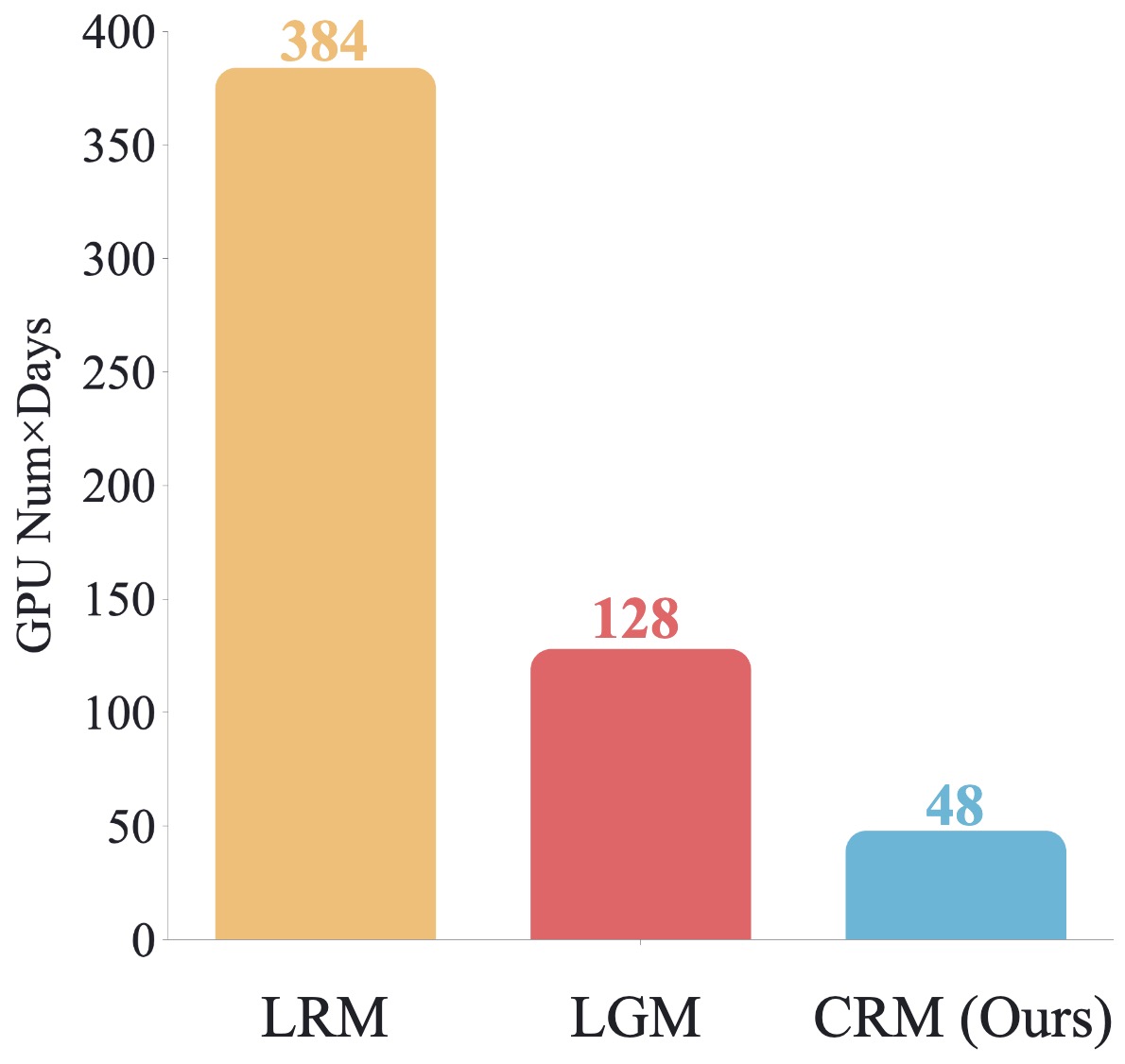}
	\caption{Training cost comparison. Our model require much less computation cost than other baselines.}
	\label{fig:training_cost}
    \end{minipage}
\end{figure}

\subsubsection{Reconstruction Results on Early Training Stage}
An advantage of our CRM is that it is easy to train. In fact, we find that CRM starts to show reasonable results at very early stage of training. The results are shown in Fig.~\ref{fig:ablation-training}. The results are good even with barely $280$ iterations (only 20 minutes of training). We conjecture that the fast convergence results from the strong geometry prior in our architecture design.
\subsubsection{Training Time of CRM}
In Fig.~\ref{fig:training_cost} we compare the training cost between our method (reconstruction model only) and two baselines, LRM~\cite{hong2023lrm} and LGM~\cite{tang2024lgm}. We measure the training cost by the training days multiplying the amount of used NVIDIA A100/A800 GPU cards. It can be seen that our model takes much smaller training time than the two baselines. This is because our model utilizes the spatial correspondence between input six orthographic images/CCMs and triplanes, which serves as a strong prior that makes the training easier.

\subsubsection{Importance of Input CCM} 
\begin{figure}[!ht]
	\centering
	\includegraphics[width=0.95\linewidth]{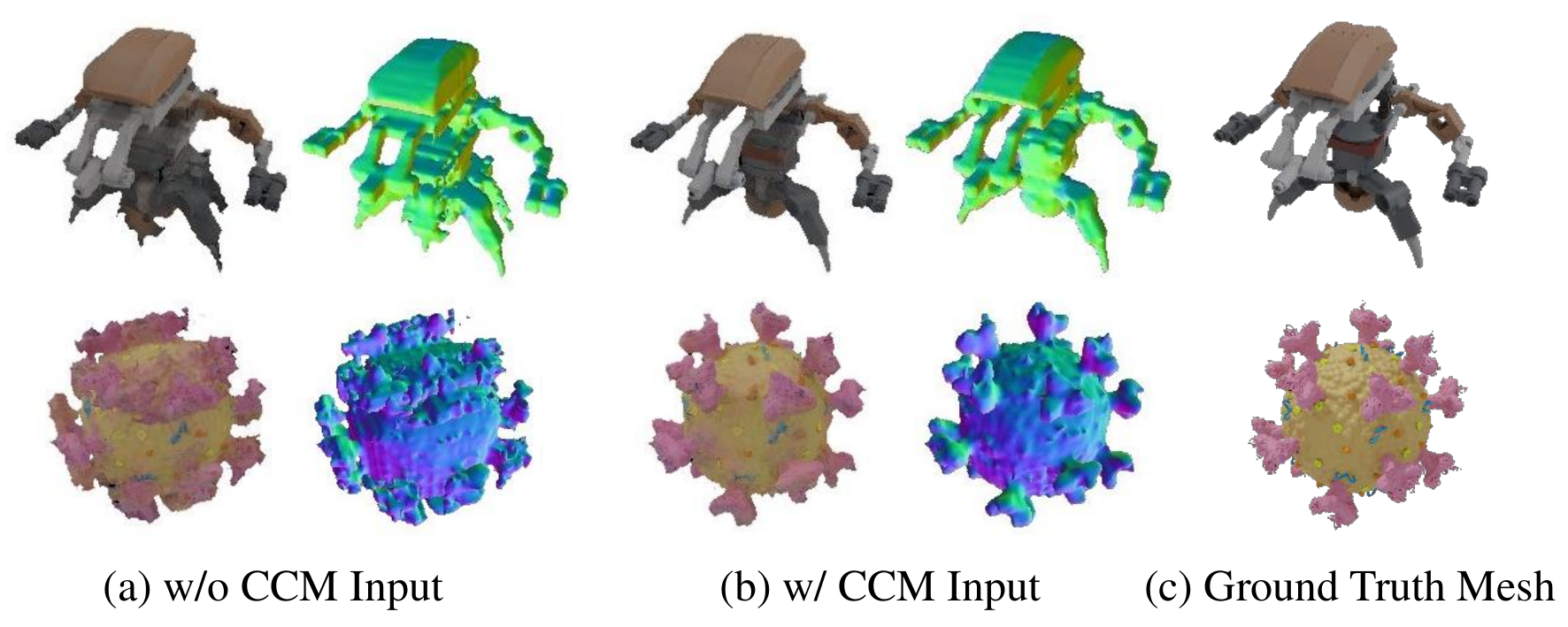}
	\caption{The CCM concatenated to the input images is beneficial for our model. (a) Without providing CCM, the model outputs a geometry which is reasonable, but not very good. (b) The shape reconstructed using our full model with CCM input, with a much better geometry. (c) Ground truth mesh rendered from the same pose.}
	\label{fig:ablation-purergb}
\end{figure}

We examine the importance of the CCMs that are concatenated to the input images. To compare, we train a reconstruction model that takes only the six RGB images as input, without CCMs. The results are shown in Fig.~\ref{fig:ablation-purergb}. It can be seen that the results of the geometry degrade a lot without CCM input. This is because CCM provides important geometry information for the model, especially when the geometry is complex.  

\subsubsection{Design of Multi-view Diffusion}
\begin{figure}[tb]
  \centering
  \includegraphics[width=0.99\linewidth]{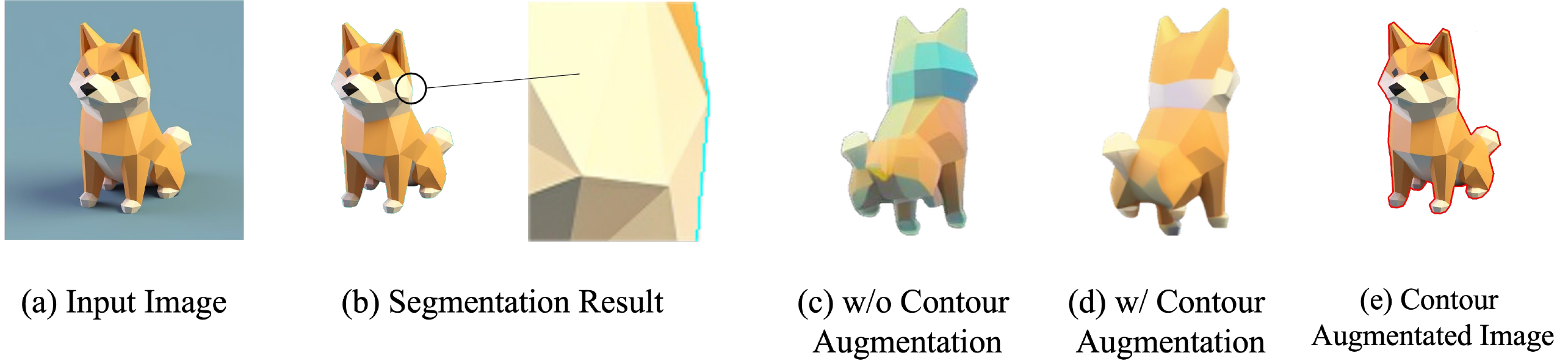}
  \caption{Demonstration of contour augmentation. (a) Given an input image, (b) off-the-shelf segmentation model sometimes provides imperfect results. (c) Without contour augmentation, the predicted backview color is sensitive to the contour. (d) With contour augmentation, the model predicts reasonable result. (e) We demonstrate how we augment the input image during training. 
  }
  \label{fig:contour_aug}
\end{figure}

Here we examine the effectiveness of the design of the multi-view diffusion models. Starting from the baseline that naively fine-tunes the pre-trained ImageDream model with 2 additional views, we sequentially add the the proposed techniques on the training. We examine the results on a subset of GSO, comparing the similarity of the generated novel view images with the ground truth images using PSNR, SSIM and LPIPS metrics. The results are shown in table~\ref{tab:ablation-diffusion}. It can be seen that both the Zero-SNR trick and random resizing are beneficial. Note that the contour augmentation does not improve the quantitative metrics. However, we find that this trick makes the model more robust to in the wild input images (Fig.~\ref{fig:contour_aug}).

\begin{table}[tb]
  \caption{Ablation study on the design of multi-view diffusion on novel view synthesis. We report the metrics of PSNR, SSIM and LPIPS on GSO dataset.}
  \label{tab:ablation-diffusion}
  \centering
  \begin{tabular}{@{}lccc@{}}
    \toprule
    Method & PSNR $\uparrow$& SSIM $\uparrow$& LPIPS$\downarrow$ \\
    \midrule
    ImageDream (6 view)     &28.99&0.8565& 0.1497\\
    + Zero-SNR              &29.13&0.8598& 0.1498\\
    + Random Resizing         &\textbf{29.36}&\textbf{0.8721}& \textbf{0.1354}\\
    + Contour Augmentation  &28.92&0.8681& 0.1444\\  
  \bottomrule
  \end{tabular}
\end{table}

\section{Conclusion}
\label{sec:conclusion}
In this work, we present a convolutional reconstruction model (CRM) for creating high-quality 3D models from a single image. Our approach effectively utilizes the spatial relationship between input images and the output triplane, leading to improved textured meshes, with significantly less training cost compared to previous transformer-based methods~\cite{hong2023lrm}. The model operates on an end-to-end training basis, directly outputting textured meshes. Overall, our method can produce detailed textured meshes in just 10 seconds.

\textbf{Limitations.} Although our method can generate a high-fidelity textured mesh in $10$ seconds, there are still some limitations. As also a limitation in ImageDream~\cite{wang2023imagedream}, if the input image has a large elevation or different FoV, the results are sometimes not satisfactory. Also, it is very hard to ensure that the multi-view diffusion model always generates fully consistent results, and inconsistent images may make the 3D results degrade. Finally, the Flexicubes grid size is only $80$ owing to the limited computing resource, which cannot represent very detailed geometry. 

\textbf{Potential Negative Impact.} Similar to many other generated models, our CRM may be used to generate malicious or fake 3D contents, which may need additional caution.


%
%
\bibliographystyle{splncs04}
\bibliography{main}

\end{document}